\documentclass[runningheads]{llncs}
\usepackage{graphicx}
\usepackage{amssymb}
\usepackage{subfigure}
\usepackage{cite}
\usepackage{booktabs}
\usepackage{bm}
\usepackage[marginal]{footmisc}
\begin{document}

\title{MVP-Net: Multi-view FPN with Position-aware Attention for Deep Universal Lesion Detection}
% \author{Zihao Li$^{1,4*}$, Shu Zhang$^{2}$\thanks{ indicates equal contribution. This work is done when Zihao Li is an intern at DeepWise AI Lab.}, Junge Zhang$^1$, Kaiqi Huang$^1$, 
% Yizhou Wang$^{2,3,4}$, Yizhou Yu$^4$
% }
% \institute{{Institute of Automation, Chinese Academy of Sciences}
% \and {Computer Science Dept., Peking University}
% \and {Peng Cheng Laboratory}
% \and {DeepWise AI Lab}
% }
\author{Zihao Li$^{1,2*}$ 
\quad Shu Zhang$^{3}$\thanks{ indicates equal contribution. This work is done when Zihao Li is an intern at Deepwise AI Lab.}
\quad Junge Zhang$^1$
\quad Kaiqi Huang$^1$\quad \newline
Yizhou Wang$^{3,2,4}$\quad 
Yizhou Yu$^2$ \newline
$^1$Institute of Automation, Chinese Academy of Sciences \quad $^2$Deepwise AI Lab\newline
\quad $^3$Computer Science Dept., Peking University \quad $^4$Peng Cheng Laboratory 
}
% index{Li, Zihao}
% index{Zhang, Shu}
% index{Zhang, Junge}
% index{Huang, Kaiqi}
% index{Wang, Yizhou}
% index{Yu, Yizhou}
\institute{}
\authorrunning{Z. Li et al.}
\titlerunning{MVP-Net: Multi-view FPN with Position-aware Attention}

\maketitle

\begin{abstract}
Universal lesion detection (ULD) on computed tomography (CT) images is an important but underdeveloped problem. Recently, deep learning-based approaches have been proposed for ULD, aiming to learn representative features from annotated CT data. However, the hunger for data of deep learning models and the scarcity of medical annotation hinders these approaches to advance further. In this paper, we propose to incorporate domain knowledge in clinical practice into the model design of universal lesion detectors. Specifically, as radiologists tend to inspect multiple windows for an accurate diagnosis, we explicitly model this process and propose a multi-view feature pyramid network (FPN), where multi-view features are extracted from images rendered with varied window widths and window levels; to effectively combine this multi-view information, we further propose a position-aware attention module. With the proposed model design, the data-hunger problem is relieved as the learning task is made easier with the correctly induced clinical practice prior. We show promising results with the proposed model, achieving an absolute gain of $\mathbf{5.65\%}$ (in the sensitivity of FPs@4.0) over the previous state-of-the-art on the NIH DeepLesion dataset.\footnotemark[1]

\footnotetext[1]{Code is available at https://github.com/urmagicsmine/MVP-Net.}

\end{abstract}

\section{Introduction}\label{intro}

Automated detection of lesions from computed tomography (CT) scans can significantly boost the accuracy and efficiency of clinical diagnosis and disease screening. However, existing computer aided diagnosis (CAD) systems usually focus on certain types of lesions, e.g. lung nodules~\cite{APND}, focal liver lesions~\cite{Liver}, thus their clinical usage is limited. Therefore, a Universal Lesion Detector which can identify and localize different types of lesions across the whole body all at once is in urgent need.

\begin{figure}[htbp]
\centering
\subfigure[\lbrack1024, 4096\rbrack]{
\begin{minipage}[t]{0.20\linewidth}
\centering
\includegraphics[width=1in]{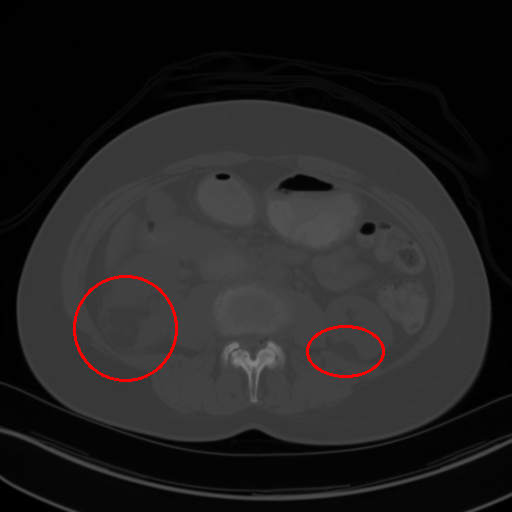}
\end{minipage} }
\subfigure[\lbrack50, 449\rbrack]{
\begin{minipage}[t]{0.20\linewidth}
\centering
\includegraphics[width=1in]{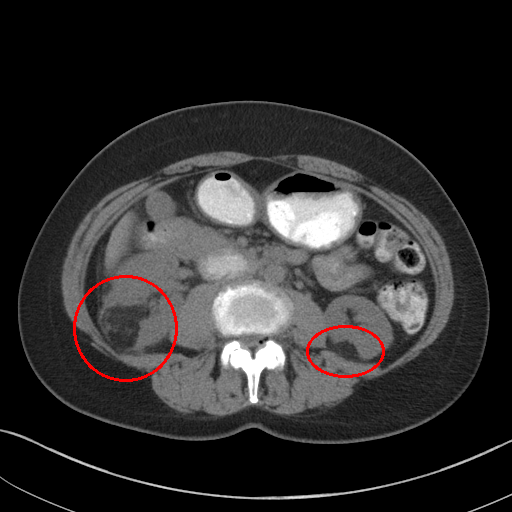}
\end{minipage}}
\subfigure[\lbrack-505, 1980\rbrack]{
\begin{minipage}[t]{0.20\linewidth}
\centering
\includegraphics[width=1in]{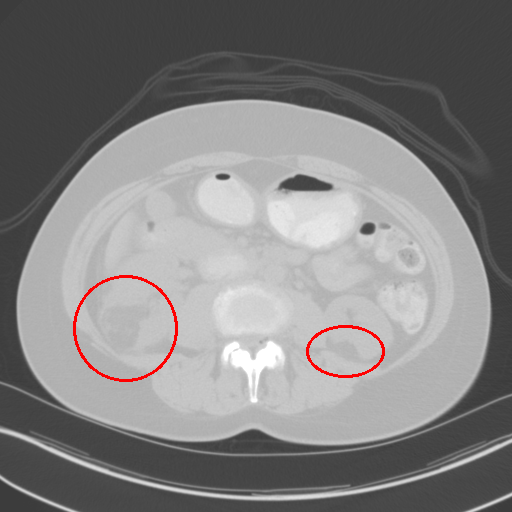}
\end{minipage}}
\subfigure[\lbrack446, 1960\rbrack]{
\begin{minipage}[t]{0.20\linewidth}
\centering
\includegraphics[width=1in]{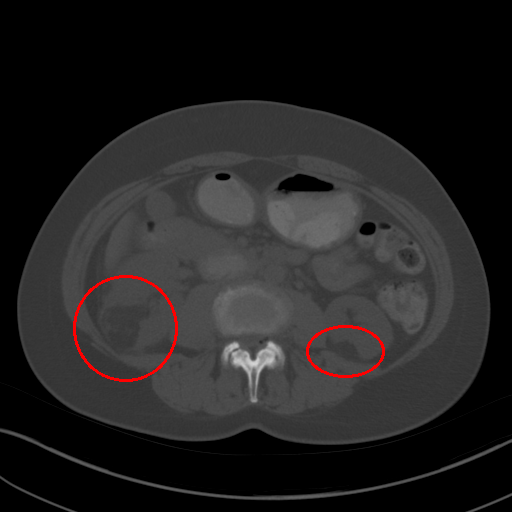}
\end{minipage}
}
\centering
\caption{CT images under different window level and window width. (a) is the image used in 3DCE. (b),(c),(d) are the multi-view images used in our MVP-Net. } \label{fig1}
\end{figure}

Previous methods for ULD are largely inspired by the successful deep models in the field of natural images. For instance, Tang \textit{et al}.~\cite{ULDOR} adapted a Mask-RCNN~\cite{MRCNN} based approach to exploit the auxiliary supervision from manually generated pseudo mask. On the other hand, Yan \textit{et al}. proposed a 3D Context Enhanced (3DCE) RCNN model~\cite{3DCE} which harness ImageNet pre-trained models for 3D context modeling. Due to a certain degree of resemblance between natural images and CT images, these advanced deep architectures % , which have achieved superior performance on natural images, %also demonstrated outstanding ability in addressing the ULD problem. 
also demonstrated impressive results on ULD.

Nonetheless, the intrinsic quality of medical images should not be overlooked. Beyond that, the inspection of medical images also exhibits different characteristics compared with recognition and detection of natural images. Therefore, it would be helpful if we can efficiently exploit proper domain knowledge to develop deep learning based diagnosis systems. We will try to analyze two aspects of such domain knowledge, and explore how to formulate these human expertise into a unified deep learning framework.

To accommodate for network input, previous studies~\cite{ULDOR, 3DCE} use a significantly wide window\footnotemark[2] to compress CT's 12bit Hounsfield Uint (HU). However, this would severely deteriorate the visibility of lesions as a result of degenerated image contrast, as shown in Fig.\ref{fig1}(a). In the clinical practice, fusing information from multiple windows are effective in improving the accuracy of detecting subtle lesions and reducing false positives (FPs). During visual inspection of the CT images, radiologists would combine complex information of different inner structures and tissues from multiple reconstructions under different window widths and window levels to locate possible lesions. To imitate this process, we propose to extract prominent features from three frequently examined window widths and window levels and capture complementary information across different windows with an attention based feature aggregation module. 
% \footnote{Windowing, also known as gray-level mapping, is used to change the appearance of the picture to highlight particular structures.}

\footnotetext[2]{Windowing, also known as gray-level mapping, is used to change the appearance of the picture to highlight particular structures.}

During the inspection of whole body CT, body position of a slice (i.e. the z-axis position of a certain slice), is also a frequently consulted prior knowledge. Experienced specialists often rely on the underlying correspondence between body position and lesion types to conduct lesion diagnosis. Moreover, radiologists would use the position cue as an indicator for choosing proper window width and window level. %besides the appearance of the CT slice, which reflects the density of body tissues in that slice, the position cue is also an indicator for choosing proper window width and window level. 
For instance, radiologists will mainly refer to the lung, bone and mediastinal window when inspecting a chest CT. Therefore, it would be very beneficial if we can exploit the position information to conduct lesion diagnosis and window selection in designing our deep detector.

\begin{figure}[!tbp]
\includegraphics[width=\textwidth]{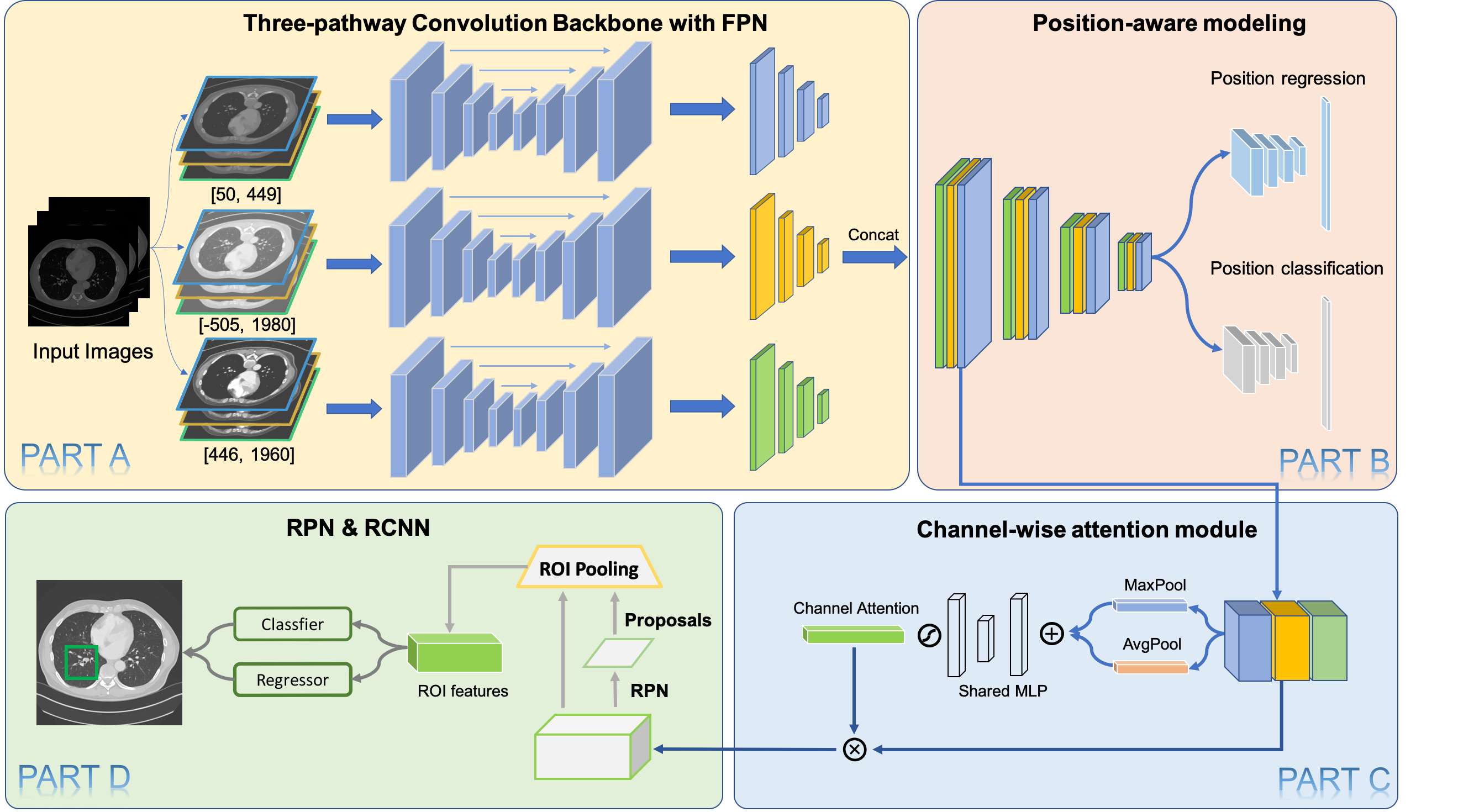}
\caption{Overview of our proposed MVP-Net. Coarser feature maps of FPN are omitted in part C and D for clarity, they use the same attention module with shared parameters for feature aggregation.} \label{fig2}
\end{figure}

In order to model the aforementioned domain knowledge and human expertise, we develop a MVP-Net (Multi-View FPN with Position-aware attention) for universal lesion detection. FPN~\cite{FPN} is used as a building block to improve detection performance for small lesions. To leverage information from multiple window reconstructions, we build a multi-view FPN to extract multi-view\footnotemark[3] features using a three-pathway architecture. Then, an channel-wise attention module is employed to capture complementary information across different views. To further consider the position cues, we develop a multi-task learning scheme to embed the position information onto the appearance features. Thus, we can explicitly condition the lesion finding problem on the entangled appearance and position features. Moreover, by connecting the proposed attention module to such an entangled feature, we are able to conduct position-aware feature aggregations. Extensive experiments on the DeepLesion Dataset validate the effectiveness of our proposed MVP-Net. We can achieve an absolute gain of $\mathbf{5.65\%}$ over the previous state-of-the-arts (3DCE with 27 slices) while considering 3D context from only 9 slices.

% \footnote{As a common practice in machine learning, we refer to reconstruction under a certain window width and window level as a view of that CT.}
\footnotetext[3]{As a common practice in machine learning, we refer to reconstruction under a certain window width and window level as a view of that CT.}

\section{Methodology}

Fig.\ref{fig2} gives an overview of the MVP-Net. For simplicity, we illustrate the case that take three consecutive CT slices as network input. It should be noted that MVP-Net can be easily extended to alternatives that take multiple slices as input to consider 3D context like 3DCE~\cite{3DCE}. 

The proposed MVP-Net takes three views of the original CT scan as input and employs a late fusion strategy to fuse multi-view features before region proposal network (RPN). As shown in part A of Fig.\ref{fig2}, multi-view features are extracted from the three pathway backbone with shared parameters. Then, in part B, to exploit the position information, a position recognition network is attached to the concatenated multi-view features before RPN. Finally, a position-aware attention module is further introduced to aggregate the multi-view features, which will be passed to the RPN and RCNN network for the final lesion detection. We will elaborate these building blocks in the following subsections.

\subsection{Multi-view FPN}

The multi-view input for the MVP-Net is composed of multiple reconstructions under different window widths and window levels. Specifically, we adopt k-means algorithm to cluster the recommended windows (labeled by radiologists) in the DeepLesion dataset and obtain three most frequently inspected windows, whose window levels and window widths are $\lbrack50, 449\rbrack$, $\lbrack-505, 1980\rbrack$ and $\lbrack446, 1960\rbrack$ respectively. As shown in Fig.\ref{fig1}, these clustered windows approximately correspond to the soft-tissue window, lung window, and the union of bone, brain, and mediastinal windows respectively.

As shown in Fig.\ref{fig2}, we adopt a three pathway architecture to extract the most prominent features from each representative view. %Backbone of each pathway is a FPN~\cite{FPN} in our implementation. To take 3D context into consideration, a 3-channel image generated by grouping three consecutive slices together is used as the input for each pathway. 
FPN~\cite{FPN} is used as the backbone network of each pathway. It takes in three consecutive slices as input to model 3D context.%, stacking them to a 3-channel image before feeding into the first convolution layer.

\subsection{Attention based Feature Aggregation}

Features extracted from different views (windows) need to be properly aggregated for accurate diagnosis. A naive implementation for feature aggregation could be concatenating them along the channel dimension. However, such an implementation would have to rely on the following convolution layers for effective feature fusion.

% In the proposed MVP-Net, we employ an attention module with a combination of channel-wise and spatial-wise attention to explicitly model the feature selection and fusion process. The channel-wise attention module can be used to adaptively reweight the feature maps of different views, imitating the process that radiologists put different weights on multiple views for lesion identification. 

In the proposed MVP-Net, we employ a channel-wise attention based feature aggregation mechanism to adaptively reweight the feature maps of different views, imitating the process that radiologists put different weights on multiple windows for lesion identification. We adopt an implementation similar to the Convolutional Block Attention Module (CBAM)~\cite{CBAM} to realize the channel-wise attention. Details for the attention module is shown in Fig.\ref{fig2}. Denoting \bm{$F$} as input feature map, we firstly aggregate the features with average pooling $P_{avg}$ and max pooling $P_{max}$ separately to extract representative descriptions, then a fully-connected bottleneck module \bm{$\theta(\cdot)$} and a sigmoid layer \bm{$\sigma(\cdot)$} are sequentially applied to the aggregated features to generate combinational weights of different channels. Multiplying \bm{$F$} by the weights, the output \bm{$F_c$} of the feature aggregation module is can be described as Eq.\ref{equ:attention}:

\begin{equation}\label{equ:attention}
\bm{F_c} = \bm{F}\cdot\sigma(\theta(P_{avg}(\bm{F})+ P_{max}(\bm{F}))).
\end{equation}

%While the spatial-wise attention mimic the process that lesions from different spatial positions are inspected with varied views. 

\subsection{Position-aware Modeling}

Due to FPN's large receptive field, the position information in the xy plane  (or context information) has already been inherently modeled. Therefore, we mainly focus on the modeling of the position information along the z-axis. Specifically, we propose to learn position-aware appearance features by introducing a position prediction task during training. Entangled position and appearance features are learned through the multi-task design in the MVP-Net that jointly predicts the position and the detection bounding box.

% \begin{equation}
% \mathbf{\mathcal{L}_{pos}}  = -\frac{1}{n}\sum^N_i y \ln p_i + (1-y)\ln (1-p_i) + \frac{1}{n}\sum^N_i\left(t - t^*_i\textbf{}\right)^2
% \label{eq1}
% \end{equation}

\begin{equation}\label{equ:position}
\mathbf{\mathcal{L}_{pos}}  = -\frac{1}{n}\sum^n_i y_i \log \phi(x_i) + \frac{1}{n}\sum^n_i\left(p_i - \psi(x_i)\textbf{}\right)^2
\end{equation}

As shown in Fig.\ref{fig2}, our position prediction module is supervised by two losses: a regression loss and a classification loss. The regression loss is applied after the continuous position regressor, whose learning target are generated by a self-supervised body-part regressor~\cite{UBPR} in the DeepLesion Dataset\cite{DeepLesion}. Due to noise in the continuous labels, we further divide position values into three classes (chest, abdomen, and pelvis) according to the distribution of the position values on the z-axis, and use a classification loss to learn this discrete position, as it is more robust to noise and improves training stability.

%As shown in Fig.\ref{fig2}, we employ a joint regression and classification loss for position prediction. In the DeepLesion Dataset\cite{DeepLesion}, continuous labels for the z-axis position was generated by a self-supervised body-part regressor~\cite{UBPR}. However, there exists a certain degree of noise in the continuous labels. To remedy this issue and bring stability to the training process, we manually divide position values into three discrete classes, i.e. chest, abdomen, and pelvis, according to the distribution of z-axis. 
%We employ the cross entropy loss and the euclidean loss for the two tasks respectively.
Let \bm{$y,p$} denote the ground-truth of discrete and continuous position values, given the bottleneck feature \bm{$x$} of FPN, we use two subnets \bm{$\phi(\cdot)$}, \bm{$\psi(\cdot)$} of several CNN layers to obtain the corresponding predictions. The overall loss function of the position module is described in Eq.\ref{equ:position} .

% Let $h(\cdot)$ denote the quantitative function, given a predicted position value $p^*$ and ground-truth $p$, we employ cross entropy loss and euclidean loss
%  for the joint classification and regression task respectively, as described in equation 1.
% \begin{equation}
% \mathbf{\mathcal{L}_{pos}}  = -\frac{1}{n}\sum^N_i h(p_i) \log h(p^*_i) + \frac{1}{n}\sum^N_i\left(p - p^*_i\textbf{}\right)^2
% \label{eq1}
% \end{equation}

%The overall loss function is:
%\begin{equation}
%L = L_{bbox\_cls} + L_{bbox\_reg} + \alpha * L_{pos\_cls} + \beta * L_{pos\_reg}
%\end{equation}

\section{Experiments}

\subsection{Experimental Setup}

\noindent\textbf{Dataset and Metric} The NIH DeepLesion\cite{DeepLesion} is a large-scale CT dataset, consisting of 32,735 lesion instances on 32,120 axial CT slices. Algorithm evaluation is conducted on the official testing set ($15\%$), and we report sensitivity at various FPs per image levels as the evaluation metric. For simplicity, we mainly compare the sensitivity at 4 FPs per image in the text of the following subsections.

% froc curve
%\begin{figure}
%\includegraphics[width=\textwidth]{froc.eps}
%\caption{FROC curves of various methods on the test set of DeepLesion.} \label{fig-roc}
%\end{figure}
% FP @0.5~16
\begin{table}[!htbp]
    \caption{Sensitivity (\%) at various FPs per image on the testing set of DeepLesion. We don't provide results with 27 slices due to memory limitation. $\mathbf{^*}$ indicates re-implementation of 3DCE with FPN as backbone.}
    \label{tab1}
    \centering
    \footnotesize % fontsize
    \setlength{\tabcolsep}{8pt}% column separation
    \renewcommand{\arraystretch}{1.2}%row space 
        \centering
        \label{tab:sample_1}
        \begin{tabular}{lcccccccc}
        \toprule
            \textbf{FPs per image} & $\mathbf{0.5}$ & $\mathbf{1}$ & $\mathbf{2}$ & $\mathbf{3}$ & $\mathbf{4}$ \\
            %\cline{2-9}% partial hline from column i to column j
            \hline
            \hline
            ULDOR\cite{ULDOR} & 52.86 & 64.80 & 74.84 &-& 84.38 \\
            3DCE, 3 slices\cite{3DCE} & 55.70 & 67.26 & 75.37 &-& 82.21 \\
			3DCE, 9 slices\cite{3DCE} & 59.32 & 70.68 & 79.09 &-& 84.34 \\
			3DCE, 27 slices\cite{3DCE} & 62.48 & 73.37 & 80.70 &-& 85.65 \\
            \hline
            FPN+3DCE, 3 slices$\mathbf{^*}$ & 58.06 & 68.85 & 77.48 & 81.03 & 83.27 \\
            FPN+3DCE, 9 slices$\mathbf{^*}$  &64.25 & 74.41 & 81.90 & 85.02 & 87.21 \\
            FPN+3DCE, 27 slices$\mathbf{^*}$  & 67.32 & 76.34 & 82.90 & 85.67 & 87.60 \\
            \hline
            \textbf{Ours, 3 slices} & \textbf{70.01} & \textbf{78.77} & \textbf{84.71} & \textbf{87.58} & \textbf{89.03} \\
            \textbf{Ours, 9 slices} & \textbf{73.83} & \textbf{81.82} & \textbf{87.60} & \textbf{89.57} & \textbf{91.30} \\
            \hline
            \textbf{Imp over} 3DCE, 27slices\cite{3DCE} & $\uparrow$ \textbf{11.35} & $\uparrow$8.45 & $\uparrow$6.90 & \textbf{-} & $\uparrow$5.65 \\
        \end{tabular}
\end{table}

\noindent\textbf{Baselines} We compare our proposed MVP-Net with two state-of-the-art methods, i.e. 3DCE~\cite{3DCE} and ULDOR~\cite{ULDOR}. ULDOR adopts Mask-RCNN for improved detection performance, while 3DCE exploits 3D context to obtain superior lesion detection results. Previous best results are achieved by 3DCE when using 27 slices to model the 3D context.

\noindent\textbf{Implementation Details}  We use FPN with ResNet-50 for all experiments. Parameters of the backbone are initialized from the ImageNet pre-trained models, and all other layers are randomly initialized. Anchor scales in FPN are set to (16, 32, 64, 128, 256). We normalize the CT slices in the z-axis to a slice interval of 2 mm, and then resize them to 800 pixels in the xy-plane for both training and testing. We augment training data with horizontal flip, and no other data augmentation strategies are employed. The models are trained using stochastic gradient descent for 13 epochs. The base learning rate is 0.002, and it is reduced by a factor of 10 after the 10-th and 12-th epoch.

\subsection{Comparison with State-of-the-arts}

The comparison between our proposed model and the previous state-of-the-art methods are shown in Table.\ref{tab1}. As the original implementation of 3DCE is based on the R-FCN~\cite{RFCN}, we re-implement 3DCE with the FPN backbone for fair comparison.
The result show that with FPN as backbone, the 3DCE model achieves a performance boost of over 2\% compared to the RFCN based model. This validates the effectiveness of our choice of using FPN as the base network. 

More importantly, even using far less 3D context, our model with 3 slices for context modeling has already achieved SOTA detection results, outperforming 27-slices based RFCN and FPN 3DCE models by $3.38\%$ and $1.43\%$ respectively. When compared with 3-slices based counterpart, our model shows a superior performance gain of $6.82\%$ and $5.76\%$. This demonstrates the effectiveness of the proposed multi-view learning strategy as well as the position-aware attention module. Finally, by incorporating more 3D context, our model with 9 slices get a further performance boost and surpasses the previous SOTA by a large margin ($\mathbf{5.65\%}$ for FPs@4.0 and $\mathbf{11.35\%}$ for FPs@0.5). %It should also be noted that our best-performed model achieves a sensitivity of over $91\%$ with only 4 FPs per image, which is only possible with over 16 FPs for the previous SOTA model.

%When compared with RFCN and FPN based 3DCE which also take 3 slices as input, our model shows a superior performance gain of over $8.79\%$ and $6.68\%$. This demonstrates the effectiveness of the proposed multi-view learning strategy and position-aware attention module. 
%although using far less 3D context,

\subsection{Ablation Study}

\begin{table}[!htbp]
    \caption{Ablation study of our approach on the DeepLesion dataset.}
    \label{tab2}
    \centering
    \footnotesize % fontsize
    \setlength{\tabcolsep}{8pt}% column separation
    \renewcommand{\arraystretch}{1.2}%row space 
        \centering
        \label{tab:sample_2}
        \begin{tabular}{lcccc|cc}
        \toprule
            FPN & Multi-view & Attention & Position & 9 slices & $\mathbf{FPs@2.0}$ & $\mathbf{FPs@4.0}$ \\
            \hline
            \hline
            $\checkmark$ &  &  &  & & 77.48 & 83.27 \\
            $\checkmark$ & $\checkmark$ && & & 81.29 & 86.18 \\
            $\checkmark$ & $\checkmark$ &$\checkmark$ & & & 84.18 & 87.89 \\
            $\checkmark$ & $\checkmark$ &$\checkmark$ & $\checkmark$ & & 84.71 & 89.03 \\
            $\checkmark$ & $\checkmark$ &$\checkmark$ & $\checkmark$ & $\checkmark$ & \textbf{87.60} & \textbf{91.30} \\
            \bottomrule
        \end{tabular}
\end{table}
%Table.\ref{tab2} presents our ablation study on the test set of the DeepLesion dataset. To comprehensively investigate the effect of each proposed modules, 
We perform ablation study on four major components: multi-view modeling, attention based feature aggregation, position-aware modeling, and 3D context enhanced modeling. As shown in Table \ref{tab2}, using simple feature concatenation for feature aggregation, the multi-view FPN obtains a $2.91\%$ improvement over the FPN baseline. Further modifying the aggregation strategy to channel-wise attention accounts for another $1.71\%$ improvement. Then learning the entangled position and appearance features with position-aware modeling further brings $1.14\%$ boost on the sensitivity. Combining our proposed approach with 3D context modeling gives the best performance.

\begin{figure}[!tb]
\includegraphics[width=\textwidth]{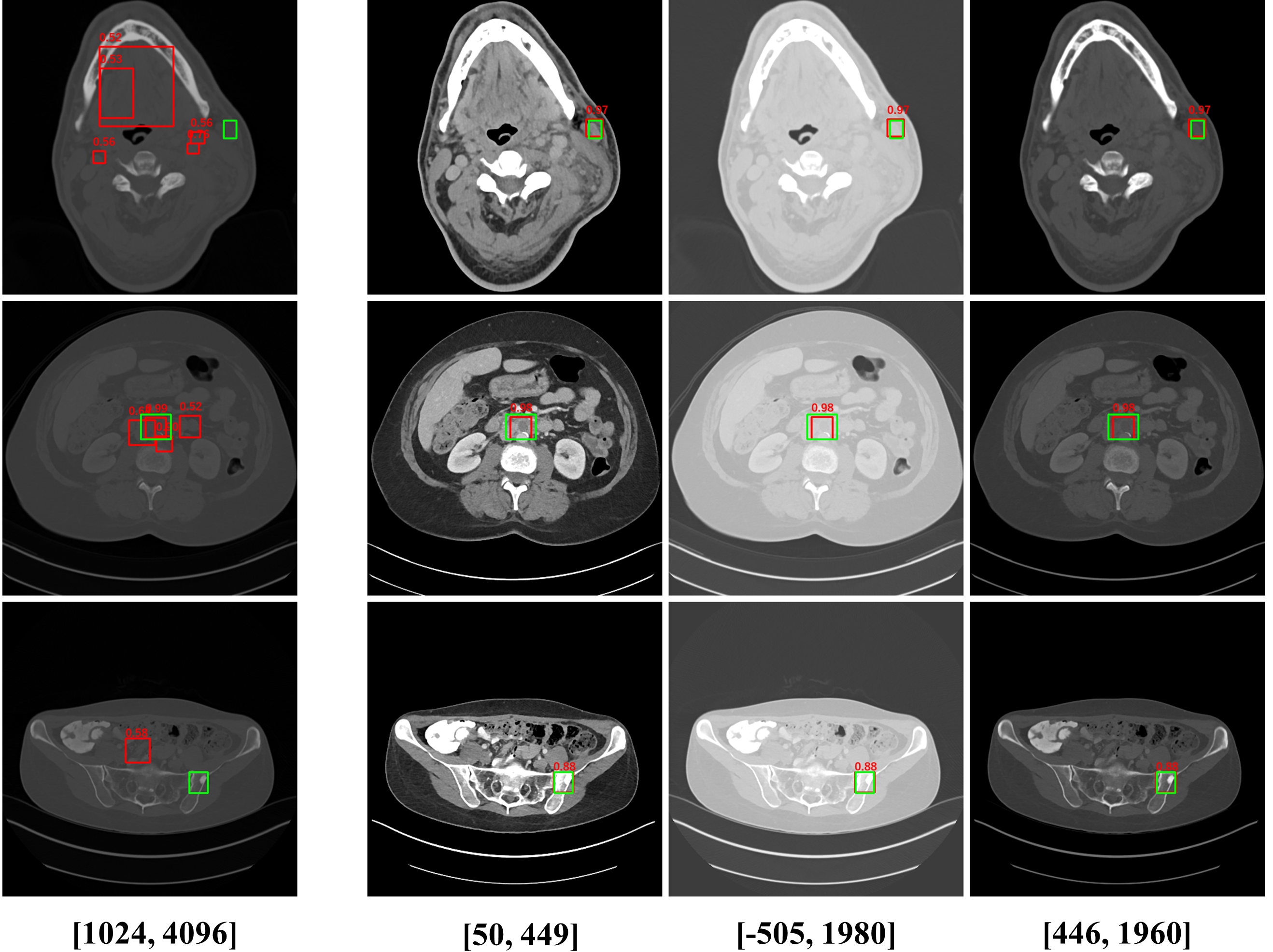}
\caption{Case study for 3DCE (left-most column) and attention based multi-view modeling (the other three columns). Green and red boxes correspond to ground-truths and predictions respectively.} \label{fig-case}
\end{figure}

We also perform a case study to analyze the importance of multi-view modeling. As shown in Fig.~\ref{fig-case}, the model indeed benefits from the multi-view modeling: the lesions that are originally indistinguishable in the view of 3DCE due to the wide window range and lack of contrast, now becomes distinguishable under the view of appropriate windows. Thus our model presents better identification and localization performance.
%our multi-view approach apparently gives better identification and localization performance because lesion and non-lesion areas have much higher contrast in our multi-view input. For instance, in case 2, the lesions can hardly be seen in the wide windowing input of 3DCE but it can be easily distinguished from our multi-view input.
%We study the effect of our multi-view modeling with a qualitative case study on Fig.\ref{fig-case}. 
%Specifically, we compare FPN based 3DCE with our MVP-Net without position modeling. Our multi-view approach apparently gives better identification and localization performance because lesion and non-lesion areas have much higher contrast in our multi-view input. For instance, in case 2, the lesions can hardly be seen in the wide windowing input of 3DCE but it can be easily distinguished from our multi-view input.

\section{Conclusion}
In this paper, we address the universal lesion detection problem by incorporating human expertise into the design of deep architecture. Specifically, 
%by observing how radiologists conduct diagnosis on the whole body CT, 
we propose a multi-view FPN with position-aware attention (MVP-Net) to incorporate the clinical practice of multi-window inspection and position-aware diagnosis to the deep detector. Without bells and whistles, our proposed model, which is intuitive and simple to implement, improves current state-of-the-art by a large margin. The MVP-Net reduces the FPs to reach a sensitivity of $91\%$ by over three quarters (from 16 to 4) and reaches a sensitivity of $87.60\%$ with only 2 FPs per image, making it more applicable to serve as an initial screening tool on daily clinical practice.

\noindent\textbf{Acknowledgement} This work is funded by the National Natural Science Foundation of China (Grant No. 61876181, 61721004, 61403383, 61625201, 61527804) and the Projects of Chinese Academy of Sciences (Grant QYZDB-SSW-JSC006 and Grant 173211KYSB20160008). We would like to thank Feng Liu for
valuable discussions.

\end{document}